\documentclass[trans]{IEEEtran}

\usepackage{times}  
\usepackage{helvet} 
\usepackage{courier}  
\usepackage{qtree} 
\usepackage{listings} 
\usepackage{xcolor}
\usepackage{amsfonts}
\usepackage{graphicx}
\usepackage{mathrsfs}
\usepackage{amsmath}
\usepackage{amsthm}
\usepackage{amscd}
\usepackage{tabularx}
\usepackage{url}
\usepackage{multirow}
\usepackage{subfigure}
\usepackage{float}
\usepackage[caption=false]{subfig}
\usepackage{array}
\usepackage{algorithm}
\usepackage{amsmath}
\usepackage{bm}
\usepackage{comment}
\usepackage{makecell}

\usepackage[justification=centering]{caption}

\usepackage{algpseudocode}

\algnewcommand\algorithmicforeach{\textbf{for each}}
\algdef{S}[FOR]{ForEach}[1]{\algorithmicforeach\ #1\ \algorithmicdo}

\title{TSDCRF: Balancing Privacy and Multi-Object Tracking via Time-Series CRF and Normalized Control Penalty}

\begin{document}

\author{\IEEEauthorblockN{Bo Ma, Jinsong Wu, ~\IEEEmembership{Senior Member~IEEE} , Weiqi Yan
\thanks{Corresponding author: Bo Ma}
\thanks{Bo Ma and Weiqi Yan are with the School of Mathematics and Computer Engineering,
Auckland University of Technology,Auckland  1024, New Zealand, email: rcn4743@aut.ac.nz. }
\thanks{Jinsong Wu is with Department of Electrical Engineering, University of Chile, Santiago, Chile, email: wujs@ieee.org}
}\\

}

\maketitle

%




\begin{abstract}
Multi-object tracking in video often requires appearance or location cues that can reveal sensitive identity information, while adding privacy-preserving noise typically disrupts cross-frame association and causes ID switches or target loss. We propose TSDCRF, a plug-in refinement framework that balances privacy and tracking by combining three components: (i) $(\varepsilon,\delta)$-differential privacy via calibrated Gaussian noise on sensitive regions under a configurable privacy budget; (ii) a Normalized Control Penalty (NCP) that down-weights unstable or conflicting class predictions before noise injection to stabilize association; and (iii) a time-series dynamic conditional random field (DCRF) that enforces temporal consistency and corrects trajectory deviation after noise, mitigating ID switches and resilience to trajectory hijacking. The pipeline is agnostic to the choice of detector and tracker (e.g., YOLOv4 and DeepSORT). We evaluate on MOT16, MOT17, Cityscapes, and KITTI. Results show that TSDCRF achieves a better privacy--utility trade-off than white noise and prior methods (NTPD, PPDTSA): lower KL-divergence shift, lower tracking RMSE, and improved robustness under trajectory hijacking while preserving privacy. Source code in https://github.com/mabo1215/TSDCRF.git
\end{abstract}



\maketitle


%

\section{Introduction}
%
%
%
%

Multi-object tracking in video relies on appearance or location cues that can leak sensitive identity information; at the same time, adding privacy-preserving noise to protect such information often disrupts cross-frame association and leads to ID switches or target loss~\cite{guo2021topology}. This conflict is acute in applications such as autonomous driving, where trajectory forgery or hijacking can mislead the system and cause safety risks. We propose TSDCRF (Time-Series Dynamic Conditional Random Field), a plug-in refinement framework that balances privacy and tracking by: (i) injecting calibrated $(\varepsilon,\delta)$-differentially private noise on sensitive regions; (ii) applying a Normalized Control Penalty (NCP) to stabilize class predictions before noise; and (iii) using a time-series CRF to enforce temporal consistency and correct trajectory deviation after noise, thereby mitigating ID switches and improving robustness to trajectory hijacking. The framework is agnostic to the underlying detector and tracker (e.g., YOLOv4 and DeepSORT). In the rest of this section we state our motivations, the threat model, limitations of standard differential privacy in tracking, and our contributions.

\paragraph{Motivations.}
Object tracking in video inherently uses appearance and motion cues that can reveal identity (e.g., face, gait, or re-identification from ROI features). Simply adding strong noise to protect privacy, however, distorts detections and breaks data association across frames, leading to ID switches and target loss. Therefore, a method is needed that (a) applies privacy noise only where necessary (e.g., sensitive classes), (b) stabilizes association by down-weighting unreliable predictions before noise (NCP), and (c) corrects trajectory deviation after noise using temporal consistency (time-series CRF). Our motivations are thus: to achieve a \emph{balance} between privacy protection and tracking utility; to make the design \emph{plug-in} so it can augment existing detectors and trackers; and to provide \emph{quantifiable} trade-offs (e.g., via privacy budget $\varepsilon$, KL-divergence, and PSNR) so practitioners can tune the system. These motivations drive the three components of TSDCRF and the experiments in Section~VI.

\begin{figure}[ht!]
	\centering
	\includegraphics[width=0.5\textwidth]{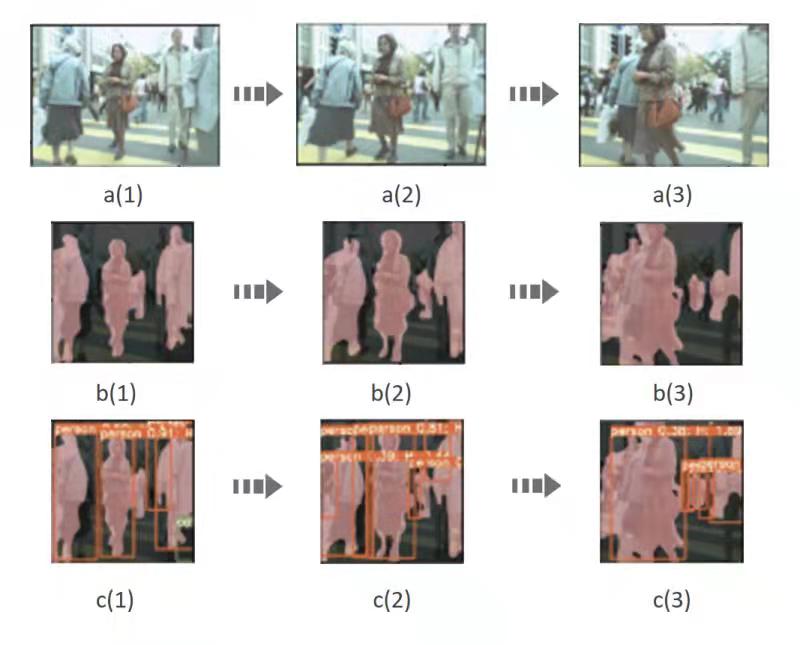}
	\caption{Segmentation and tracking results. Top: three consecutive frames from MOT16. Bottom: segmentation and trajectory after target identification.}\label{fig:tracking}
\end{figure}

The data in Fig.~\ref{fig:tracking} are from MOT16~\cite{milan2016mot16}. The three images on the \emph{top} are consecutive frames; the three images on the \emph{bottom} show the segmentation and the movement trajectory after identifying the target.

\begin{figure}[ht!]
	\centering
	\includegraphics[width=0.5\textwidth]{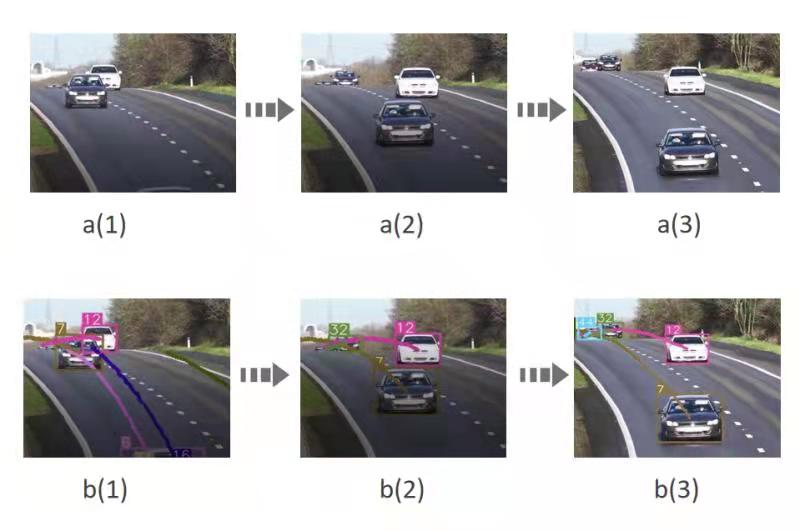}
	\caption{Tracking Result For Vehicle}\label{fig:tracking1}
\end{figure}

\begin{figure*}[ht!]
	\centering
	\includegraphics[width=0.8\textwidth]{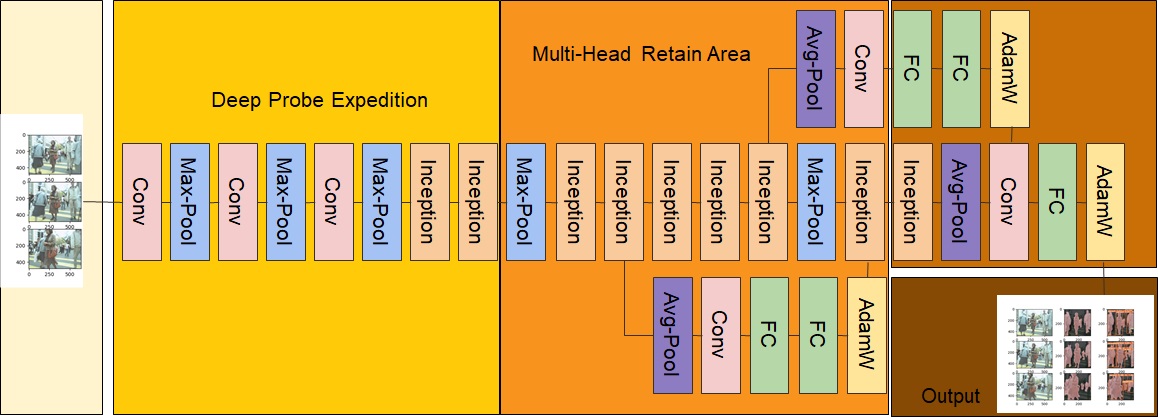}
	\caption{Flowchart of the proposed framework: detector $\rightarrow$ tracker $\rightarrow$ privacy module $\rightarrow$ TSDCRF refinement.}\label{fig:arc}
\end{figure*}

\subsection{Threat model}\label{sec:threat}
We consider an adversary who can access the tracking outputs and intermediate representations (e.g., cropped ROI features) and aims to infer or spoof sensitive identity information.

\subsubsection{Attack 1: Feature-stealing / identity spoofing}
Given a sensitive ROI (e.g., a face) and its embedding $x$, the adversary attempts to reconstruct a spoofed representation $x'$ that matches the original identity under a verifier (e.g., face authentication~\cite{du2020cross,yan2022domain}). Our goal is to ensure that after privacy perturbation, $x'$ no longer preserves sufficient identity evidence to pass verification, while tracking association remains stable.

\subsubsection{Attack 2: Track and hijack the target orbit}
We assume that the adversary predicts the trajectory of the tracked target (e.g., a car) based on the time-series flow. Initially at time $t_0$, the adversary predicts the direction of the tracked target (e.g., to the left). At time $t_1$, the adversary posts an attack patch (e.g., on the back of the car), trying to trick the detector with two effects: (1) removing the bounding box of the target from the detection result, or (2) shifting the bounding box in a direction specified by the adversary. The constructed bounding box is then associated with the original trajectory in the tracking result, hijacking the trajectory and misleading the system. In a real scenario, the vehicle may crash into other vehicles or pedestrians. This second attack is real-time. Even if Attack 1 is mitigated, the user must be given a correct detection window in the first few frames so that the tracking framework can estimate the correct trajectory. The defense against the former cannot interfere with the tracking task (i.e., cannot affect the estimation of the target window). Therefore, this paper uses a time-series algorithm to compute the state of each target node from a global perspective to correct the wrong deviation of the target window and trajectory caused by the attack.


\subsection{Limitations of Differential Privacy mechanism for object tracking}
Some researchers have applied the Gaussian distribution mechanism to remove the limitations~\cite{liu2018generalized} on the improved analysis of the statistical distribution with data and integrate this mechanism into the necessary process, others proposed Topology-Aware Differential Privacy~\cite{guo2021topology} to reduce the noise scale and improve the model usability in images recognition. Thus, the image recognition system (object tracking is one instance) can generate differential privacy protection algorithms from the region of interest (ROI), which may cover the trajectory area. However, this process still needs to be improved. In the work~\cite{zhao2020novel}, Zhao mentioned that removing or replacing individuals in the data set has a negligible effect on the output distribution, and the calculations on private data will not reveal sensitive information about the individuals in the dataset. Zhao also mentioned that differential privacy could only be directly or randomly applied in trajectory data once rationality and correctness are ensured in the data sets. 

Based on the differential privacy, Dwork proposed the Theorem of the perturbation output of $(\varepsilon,\delta)$-differential privacy in~\cite{dwork2009complexity}. Namely, for any $(\varepsilon,\delta)\in (0,1),$ the Gaussian output perturbation mechanism $\sigma \leq \frac{\Delta p_2 log\frac{1.25}{\delta}}{\varepsilon}$  is $(\varepsilon,\delta)$-differential privacy.  Thus we unfold a natural question about tracking with differential privacy in Figure~\ref{fig:tracking}: 

\begin{enumerate}
    \item Does the value of $\delta$ in $(\varepsilon,\delta)$-differential privacy provides the minimum amount of noise required to obtain $(\varepsilon,\delta)$-differential privacy with Gaussian perturbations for tracking feature in inference images?
    \item What is the performance change when $\varepsilon \geq 1$ with proposed privacy-preserving Dynamic Conditional Random Field?  
\end{enumerate}

We will unfold this topic and discuss both issues: 
\begin{itemize}
\item We try to find out the $\delta$ value for the local-optimal in the image tracking solution, i.e.,  $\varepsilon \rightarrow 0$.
\item We then tackle the differential privacy problem of corrupted images and disturbing tracking algorithm according to integrating the Gaussian mechanisms with the conditional random field. 
\item At the end, we show the large values for $\varepsilon$. The standard deviation of the Gaussian perturbation that provides $(\varepsilon,\delta)$-differential privacy must be scaled like $\Omega(\frac{1}{\sqrt{\varepsilon}})$. 
    
\end{itemize}

\subsection{Contributions}
The main contributions of this paper are as follows.

\begin{enumerate}
    \item \textbf{TSDCRF refinement.} We propose a time-series dynamic CRF (TSDCRF) that models cross-frame association as a linear-chain CRF over time slices. It enforces temporal consistency of track labels and positions after privacy noise is applied, thereby mitigating ID switches and trajectory deviation caused by sensitive-region obfuscation. This refinement is plug-in and works with any detector and tracker.
    \item \textbf{Normalized Control Penalty (NCP).} We introduce NCP to assign a penalty weight to each detection before adding privacy noise: sensitive-class detections receive higher weight (more noise for stronger protection), while low-confidence or conflicting class predictions are down-weighted so that association remains stable. NCP improves the privacy--utility trade-off and is integrated into the CRF unary term.
    \item \textbf{Systematic evaluation.} We provide an empirical study on MOT16/17, Cityscapes, and KITTI, including: (i) distribution shift (KL-divergence) between original and privacy-processed data; (ii) tracking deviation (RMSE) and statistical comparison (PSNR) between noise and original data under different privacy budgets; and (iii) robustness under trajectory hijacking and blind target detection attacks. The results show that our method achieves a better balance between privacy and tracking than white noise and prior baselines (NTPD, PPDTSA).
\end{enumerate}

To support the above, we integrate Gaussian $(\varepsilon,\delta)$-differential privacy for the privacy features and use time-series methods to adjust the estimated trajectory when the attacker disrupts the inference. There are certain limitations for the Gaussian mechanism when $\varepsilon \rightarrow 0$~\cite{balle2018improving}; implementing differential privacy into existing object tracking thus requires careful design. Because privacy noise inevitably impacts association accuracy and may cause tracking targets to be lost, we use CRF-based refinement to mitigate perturbation and recover association. The following sections detail the pipeline, the CRF formulation, and the verification metrics.

\paragraph{Paper organization.} Section~II reviews related work. Section~III describes the pipeline and the linear-chain CRF used for refinement. Section~IV presents the dynamic conditional random field and the loss/measurement definitions used in evaluation. Section~V gives the information-loss and metric formulations. Section~VI reports experiments, and Section~VII concludes.

\section{Related Works}
The goal of privacy-preserving object tracking is to detect and track targets while protecting the privacy of the tracked objects. Three sub-tasks are involved: object detection, object tracking, and privacy protection. Below we discuss related work in detail and clarify how our approach relates to and extends prior art.

\subsection{Privacy-preserving tracking and detection}
Prior work has addressed privacy in video at the detection or tracking stage, but often without jointly optimizing for both privacy and association stability. Zhou~\cite{zhou2020personal} use MTCNN to locate sensitive regions and apply anti-noise correction instead of removal; their focus is on preserving utility of the corrected signal rather than formal differential privacy. Zhao~\cite{zhao2020novel} apply differential privacy to trajectory data via prefix-tree structures, which protects aggregate trajectory statistics but does not address per-frame bounding-box or feature perturbation in a tracking pipeline. Topology-aware differential privacy~\cite{guo2021topology} reduces noise scale in image classification by exploiting topology; it improves utility for classification but does not target cross-frame association or ID consistency. In contrast, our work combines (i) calibrated $(\varepsilon,\delta)$-DP noise on sensitive ROIs under a configurable privacy budget, (ii) NCP to stabilize class predictions \emph{before} noise so that association is less affected, and (iii) a time-series CRF to maintain association and reduce trajectory deviation \emph{after} noise---so that tracking remains usable while identity-related features are protected. We also evaluate robustness under trajectory hijacking and report statistical measures (KL-divergence, PSNR, RMSE) that relate noise to privacy budget.

\subsection{Object detection for tracking}
Object detection is the front end of our pipeline; we use existing detectors (e.g., YOLO) and do not modify their architecture. For context, we briefly review representative methods. The You Only Look Once (YOLO) family~\cite{farhadi2018yolov3,redmon2016you} provides one-stage, real-time detection suitable for embedded and video applications; YOLO has evolved to incorporate multi-scale prediction and context similar to SSD. SSD~\cite{liu2016ssd} is another one-stage method that classifies predefined anchors with convolutional networks. YOLOv4~\cite{bochkovskiy2020yolov4} and later variants improve accuracy and speed. Our framework is agnostic to the choice of detector; we use YOLO in experiments for its speed and wide adoption. The key requirement is that the detector outputs bounding boxes, class scores, and class IDs so that we can apply NCP and privacy noise on sensitive classes before passing results to the tracker.

\subsection{Privacy-preserving in object detection}
In object tracking, protecting privacy at the detection stage (per-frame) is a natural first step, since global tracking privacy can be better handled if each frame's sensitive content is already protected. Zhou~\cite{zhou2020personal} proposed a privacy protection method that uses MTCNN~\cite{zhang2016joint} to locate sensitive positions and applies anti-noise correction instead of removing sensitive information, thus preserving some utility. Wang et al.~\cite{wang2021dynamic} note that low-resolution and fast-motion video remain challenging for CNNs, and that backbone networks may perform poorly when image quality is low or when pixel-level information is ambiguous. Our approach does not replace the detector; we post-process its outputs by (i) identifying sensitive classes, (ii) applying NCP weights, and (iii) adding Gaussian DP noise to sensitive bounding boxes, then refining with a time-series CRF. This keeps the detector unchanged and allows deployment with existing industrial pipelines.

\subsection{Object tracking and self-attention}
When video contains noise (e.g., from privacy perturbation), maintaining detection and association accuracy is difficult. Attention mechanisms have been used to improve robustness. Per-channel attention (e.g., Squeeze-and-Excite (SE)~\cite{hu2018squeeze}) reduces dimensionality and captures channel dependencies but can increase computation; Efficient Channel Attention (ECA)~\cite{qilong2020eca} provides a lightweight alternative. Spatial and channel attention are combined in BAM~\cite{park2018bam} and CBAM~\cite{woo2018cbam} to refine convolutional features. Transformers have been applied to multi-object tracking (e.g., TrackFormer~\cite{meinhardt2022trackformer}), which uses a transformer to query and associate trajectories. However, integrating differential privacy with transformer-based tracking is non-trivial: DP sub-sampling and noise injection can confuse similar targets after encoding. Our approach addresses this by using a time-series CRF over track labels and positions: we associate across frames with NCP-weighted unary terms and temporal pairwise terms, so that after privacy noise is added, the CRF can still correct association errors and reduce ID switches without requiring a full transformer redesign.

\subsection{Tracking object lost and occlusion}
In multi-object tracking, occlusion and temporary target loss are common; recovering the same track ID after re-appearance is essential. Wang and Wu et al.~\cite{wang2021dynamic} proposed local and global search with dynamic object templates to handle occlusion and reduce tracking loss. Their strategy is attention-guided and improves robustness when targets disappear and reappear. In our setting, privacy noise can also cause targets to be ``lost'' (e.g., displaced or mis-associated). We treat each tracked object as a node in a time-series CRF: when an object is perturbed by privacy noise, it can still be correlated with previous frames via the CRF's pairwise terms, and the penalty (NCP) helps maintain consistent labels. Thus, objects that reappear or are temporarily masked by noise can be re-associated with the same track ID. Our contribution here is the combination of NCP and CRF-based refinement specifically under privacy and attack noise, rather than only under occlusion.

\section{Trajectory integrity and consistency Check}
\subsection{Pipeline overview}
As illustrated in Fig.~\ref{fig:arc}, the proposed framework consists of: (i) an object detector (e.g., YOLOv4) that outputs bounding boxes and class scores; (ii) a baseline tracker (e.g., DeepSORT~\cite{theagarajan2020automated}) that performs data association using motion and appearance cues; (iii) a privacy module that perturbs sensitive-region features under a given privacy budget; (iv) TSDCRF refinement that models the cross-frame association as a time-series conditional random field (CRF) and corrects association errors (e.g., ID switches) induced by privacy perturbations or adversarial trajectory hijacking.

Generally, the incomplete data-sets problem occurs due to privacy noise interference. The data are polluted due to noise interference and are partially deleted by the pre-processing module. It is important to assess the block and image integrity for privacy class detection in image privacy searches. For example, if a region (e.g., a street image with details including faces) has all the features, an attacker can match a part of the features within the image. On the other hand, consistency requires identifying all relationships relevant to a given data set. However, in the training of the target detection model, it is necessary to check the integrity of the reconstructed data results in the reconstruction discrimination matrix of Fig.~\ref{fig:arc} to avoid lacking the training accuracy of the target detection model caused by the reconstructed data.

In~\cite{bertino2005information}, a set of completeness and consistency evaluation methods were proposed. Unlike other earlier relevant works, \cite{bertino2005information} considered two more important aspects: the data's relevance and the data set's structure. Inspired by \cite{bertino2005information}, this paper combines aggregated information to evaluate the consistency via amplifying the correlation between the target data set and the data quality attributes of each aggregated information. At the same time, the formal description of each attribute involved in the aggregated information is just right. The attribute reflects on the label. On this basis, this paper proposes the missing measurement criteria (ML) as follows,
\begin{eqnarray}\label{equ:CMLdet}
ML = \sum{RTree \cdot S_i\cdot IV} + \sum {RTree\cdot S_i\cdot ROW},
\end{eqnarray}
where $RTree$ is a relationship tree (a type of directed graph) using a conditional random field, in which each node $ {S_i} $ is an attribute class, $IV$ is the integrity value, and $ROW$ is the consistency value. The lack of consistency (expressed as LC) is calculated by the multiplication of the penalty constraint violations in all sanitized transactions and the weight associated with each constraint. This penalty constraint is multiplied by the constraint of each tracked object to be the weight of its penalty coefficient.
\begin{eqnarray}\label{equ:CSLdet}
\begin{aligned}
LC = \sum{RTree\cdot TLC_i \cdot nv} + \sum{RTree \cdot TLC_i \cdot wc} \\
+ \sum{RTree \cdot DLC_j \cdot nv} + \sum{RTree \cdot DLC_j \cdot wc}
\end{aligned}
\end{eqnarray}
where $nv$ represents the number of violations, $wc$ is the weight of the constraint, $ TLC_i $ describes the simple constraint class with $i$ items , and $ DLC_j $ describes the complex constraint class with $j$ items.

In the tracking task, if the tracking objects that have been lost are cleaned up and found, the last positions of these lost objects may have a better correlation and impact on the results. Here, the user can add location attributes to these IDs to check the correlation between the same ID in different frames or classify the existing missing tracking objects to find the accurate classification of the newly-appearing objects. 

According to the expected occurrence of tracking objects, specific metrics are used to measure the loss of these objects. We define a linear-chain conditional random field (CRF) over the trajectory label sequence $\mathbf{l}=\{l_t\}_{t=1}^T$ conditioned on observations $\mathbf{o}=\{o_t\}_{t=1}^T$:
\begin{eqnarray}\label{equ:Stt}
P(\mathbf{l} \mid \mathbf{o}) = \frac{1}{Z(\mathbf{o})} \exp\Big( \sum_{t=1}^T \psi_u(l_t, o_t) + \sum_{t=2}^T \psi_p(l_{t-1}, l_t, o_{t-1}, o_t) \Big),
\end{eqnarray}
where $o_t$ includes motion cues and (privacy-perturbed) appearance features at frame $t$; $\psi_u$ is the unary term (confidence / NCP-weighted class consistency), and $\psi_p$ is the pairwise term that enforces temporal smoothness and discourages ID switches. The normalization factor $Z(\mathbf{o})$ sums over all label sequences. Objects with lost tracking IDs can be re-associated by inferring under this CRF. The tracking ID condition can be calculated as
\begin{eqnarray}\label{equ:Medet}
M_e = \frac{1}{N} \sum^{k}_{i=1} (|FI_i(pv)|-|FI_i(pv')|),
\end{eqnarray}
where $N$ represents the similarity rate from the existing ROI, $k$ is the number of searches analyzed,  $|FI_i(pv)|$ represents the number of frequent data sets for privacy, and $|FI_i(pv')|$  is the number of frequently purified data-sets for privacy, and the purified data set as $pv'$. Since privacy protection technologies usually modify data for cleaning purposes, the parameters involved in search analysis are almost inevitably affected. Therefore, to measure the privacy level of detected objects from predefined sensitive (privacy) classes, it is desired to use classified results with the sensitive classes consistently before and after applying privacy noise generation and insertion techniques.

When we need to quantify the sequence images with their IL (information loss) value, it is helpful to indicate the percentage of missing parts that is a non-sensitive pattern (i.e., association, classification rule) that is a side effect of the hiding process. In addition, the extracted information indicates the extra manual information that protects the percentage of manual modes. In~\cite{oliveira2002privacy}, Oliveira defined two measures to detect the missed costs via manual mode, and the missed costs correspond to the missing and manual information, respectively. In particular, leakage costs include the increased difficulty of processing hidden unrestricted blocks caused by privacy noise during image cleaning. By cleaning the privacy noise, we can measure the cost of failure (CF) for protecting privacy leakage by comparing the events of target data processing before and after cleaning. The mean cost with privacy leakage is given by
\begin{eqnarray}\label{equ:MCdet}
MC = \frac{T~R_p(D)-T~R_p(D')}{T~R_p(D)},
\end{eqnarray}
where $T~R_p(D)$ and $T~R_p(D')$ represent the original data set $D$ and purified data set $D'$ respectively. In the best case, the loss rate is close to 0\%. There is a trade-off between the probability of privacy protection being attacked and the object tracking accuracy rate.

The cost of failure (CF) is calculated as
\begin{eqnarray}\label{equ:MCdet2}
CF = \frac{ \sum{RTree \cdot R_p(pv)}- \sum{RTree \cdot R_p(pv')}}{ \sum{RTree \cdot R_p(pv)}},
\end{eqnarray}
where $\sum RTree \cdot R_p(pv)$ and $\sum RTree \cdot R_p(pv')$ are the unrestricted-pattern counts (with the conditional random field) for the original vectors $pv$ and the purified vectors $pv'$, respectively. In the best case, the cost of failure is 0\%. Average precision (AP) is
\begin{eqnarray}\label{equ:APdet}
AP = \frac{\Sigma|P|}{R_p (pv')},
\end{eqnarray}
where $\Sigma|P|$ is the summation of detection accuracy $P$, and $R_p(pv')$ is the detection frequency for the purified vector $pv'$. If the method has no artificial pattern label, $AP$ is 0.

In the task of tracking correlation, the lost tracked target can extract all the coordinate sets of the lost target from all the frames where the lost target appears before the lost frame, and establish the loss rule; all the lost rules that can track the ID trajectory form the collection of lost trajectories (the ghost rule). Predicting the future trajectory and detecting all objects around the predicted trajectory can help track the lost target in situations such as occlusion and establish the association set of the suspected missing object.

Similarly, if the purpose of tracking the lost target is to find the lost target and re-establish the association (e.g., by arranging all the detected targets in the predicted trajectory area into a spanning tree), we establish the lost rule for the lost target and evaluate its formation. The cost of all calculations can be quantitatively evaluated through the ghost association rules. These measures allow privacy protection algorithms to associate tracking targets via inferring suspected tracking targets. For $\varepsilon$-differential privacy, the tracked bounding box tuples $T$ have the same structure as the tracking results without privacy preservation.

\section{Dynamic Conditional Random Field and Loss Function Measurement} \label{informationlossmodel}
\paragraph{Dynamic conditional random field (DCRF).} Our refinement step is built on a dynamic conditional random field (DCRF): a linear-chain CRF over time slices~\cite{sutton2004dynamic}, where each node is a frame and the state is the track label. DCRF enforces temporal consistency of labels and positions, so that after we inject privacy noise (which perturbs locations and can break association), we can still re-associate lost tracks and correct trajectory deviation. This is why the CRF mitigates perturbation and helps track lost objects: the pairwise terms $\psi_p$ discourage ID switches and the unary terms $\psi_u$ (with NCP-weighted confidence) down-weight unreliable detections. Below we define the loss and measurement used in the experiments.

Motion trajectory estimation needs to be established so that the target window covers the area of the tracked object relatively accurately in consecutive frames. If the target window accuracy and coverage position of consecutive frames are accurate, the tracking accuracy will be improved accordingly. Before discussing tracking accuracy and what kind of data can be used to measure and protect tracking trajectories, this paper first introduces a concept, namely data accuracy. Data accuracy is closely related to data distortion caused by privacy protection: the less data distortion, the less accuracy deviation, and the better the data quality is. Data accuracy depends highly on the privacy categories that are protected during the tracking process. For improving data accuracy, we need to analyze the accuracy of these privacy protection algorithms based on different modified technologies. For the measurement of information loss, if the techniques utilize perturbation or blocking to hide the original data or clustered data, it can be found that the differences between the original data $D$ and the sanitized data $D'$, in addition, the information loss value can be calculated from the differences between  $D$ and  $D'$.

\section{Dynamic Conditional Random Field}
Dynamic conditional random field (DCRF)~\cite{sutton2004dynamic} is a conditional random field (CRF) of linear chain nature. Each node in the chain represents a time slice, and a time slice contains a set of state variables and corresponding edges and corresponding sums Parameters are bound across slices. In the task based on target tracking, the tracked target can be classified into a series of slice tasks, so the performance of DCRF in tracking is better than that of a series of linear chain CRF, and the number of parameters and cost can be halved.  Assuming that the distribution of a detection target at time $t$ is $D$, on the target trajectory vector $(A, A')$, it can be written as:
\begin{eqnarray} \label{equ:dcrfa}
D(A'|A) = \frac{1}{Z(A)} \sum_t \sum_{i\in I} exp (\Sigma_i  \epsilon_i f_i (A'_{t,I}x,t))
\end{eqnarray}
where $D(A) $ is the function to divide the time slide $t$. The features of the tracked target on DCRF form a feature set $f_i (A'_{t,I}x,t)$. On the feature set, an index can be established according to the time $t$, and the index on its feature set Both establish corresponding features $f_i $ and weights $(x,t)$. By establishing an index for each state, the tracked target is tracked by establishing a DCRF in each frame in which the target exists, and its feature $f_i $ corresponds to the confirmed detection object corresponding with each other.

\subsection{Information loss for model}

In \cite{tai2009reversible}, the authors measured the difference between the original image and the processed image through a pixel loss sampling method, which was then used in their histogram modification method. After they formed structured data by constructing the pixels of the image, they compared the frequency histograms of the unprocessed and processed images to obtain loss information.
Inspired by \cite{tai2009reversible}, we also use a histogram-based measure of information loss rate to measure the loss caused by adding privacy noise, calculated as follows:
\begin{eqnarray} \label{equ:lossrate}
LossRate(A,A')= \frac{\sum_{i=1}^n |f_A (i)-f_A' (i)|}{ \sum_{i=1}^n f_A (i)},
\end{eqnarray}
where $i$ is the data item in the original data sets $A$, and $ f_A (i) $ is its frequency in the input data sets, and $i$ is given after the application of privacy protection technology. For the $i$-th data item, $ f_A '(i) $ is the frequency of its new appearance in the data sets $A$ after privacy protection processing. 
The information loss is defined as the summation of the absolute errors between the frequency of occurrence of the data set containing no privacy noise data and the data composed of the calculated target region, and the matrix formed by the summation of errors can be formed with the frequency of the target region in the original data set. The ratio measures the proportion of errors in all data. Equation~\eqref{equ:lossrate} can also be used for algorithms to calculate the fraction of privacy-preserving data that use blocking techniques to insert uncertainty about certain sensitive data items or their dependencies into the dataset. The occurrence frequency of the sensitive data item $i$ can be used as a measure of the dataset $D'$ that does not contain privacy noises.
The minimum frequency of this sensitive data item $i$ is obtained via averaging the maximum frequency of occurrences of sensitive classes in the object detection task.

In the case of the exchange of information, the loss or cost of information caused by the heuristic algorithm can be evaluated by measured parameters, which are introduced in the exchange of values for obfuscation. If there are no correlations between different data-sets records, you can estimate the data confusions through the percentage of value replacement performed to hide specific information.

From the aspect of the exchange of entropy, the cost or loss of entropy caused by the heuristic algorithm can be evaluated via measuring parameters introduced in the exchange of values for obfuscation. For example, we assume that there are no correlations among the region of interest (ROI) corresponding to the objects in different tracked frames. In that case, the data confusion caused by external interferences can be estimated by replacing their corresponding correlation values to protect the privacy of specific objects. For multiplicative noises~\cite{liu2006random}, the size of the random projection matrix reflects the quality of the disturbance data. 
When the average value of the error boundary of the inner product matrix formed by perturbation interference approaches 0, the reciprocal of the spatial dimension observed in the eigenvector can be used as the variance. When the dimension of the random projection matrix is close to that of the original image data, the matrix's inner product generated by the image data transformation or projection is also closer to its actual value, and the error is negligible. Since the inner product here is closely related to the distance measures used in tracking methods such as DeepSORT, Euclidean distance, and Hungarian distance, and to the relationship between the feature and the processed noise image, if we directly analyze the tracking error it may bias the matching results of tracked objects between different frames. If effective processing is carried out in the early stage, these deviations can be controlled in the measured value of the distance instead of reflecting the final tracking result.

%

If modifying the region of interest (ROI) in images includes aggregating some feature points in ROI, the loss of information is given by the result of the loss in the data. Intuitively, in this case, to create the relationship between different frames, the same feature vectors extracted from different frames will utilize a "generalization or aggregation scheme" to find out the same objects in different frames, which can be ideally modeled as a tree scheme. Each cell modification applied during the clean-up phase using the general tree introduces data perturbations that reduce the general accuracy of the data sets. As in the $k$ anonymous algorithm proposed in~\cite{sweeney2002achieving}, \eqref{ILformula} can be used to sort and protect the feature vectors for the proposed privacy protection. For example, for a dataset $T$ with $N_A$ fields and $N$ transactions, if we identify the domain generalized graded $GT$ with depth $h$ as a generalization scheme, we can measure the cleaned dataset $T^*$. The information loss (IL) is given by
\begin{eqnarray}\label{ILformula}
Information Loss(T^*)=\frac{\sum^{i=N_A}_{i=l} \sum^{i=N}_{j=l} \frac{h}{|GT_{A_i}|} }{|T|*|N_A|},
\end{eqnarray}
where $\frac{h}{|GT_{A_i}|}$ represents the loss of each paired feature vector, which means that the feature vector from one frame has to match the feature vector in another frame. For concealment techniques based on sampling methods, the quality is obviously related to the sample size under consideration, generally speaking, to its feature.

\subsection{Information loss matrix}
The motion tracking algorithm uses the cost function to measure the motion window generated by several time series frames and estimate the motion trajectory of the tracking target. At the same time, in the pre-task target detection, the IOU is used to measure the similarity of the generated target windows and select the most accurate window. Therefore, to prevent the trajectory hijacking mentioned in the threat model, we need to re-introduce a set of cost measurement methods to defend against the attacker's tampering with the motion trajectory. 

We are inspired by the general loss measure (LM)~\cite{iyengar2002transforming}. We assume that $T$ is the bounding box set with $n$ as its confidence score or attributes. The LM metric is the average information loss of all data units of a given data set, defined as
\begin{eqnarray}\label{LMformula}
LM(T^*)=\frac{\sum^{n}_{i=l} \sum^{|T|}_{j=l} \frac{f(T*[i][j])-1}{G(A_i)-1} }{|T|*|N|},
\end{eqnarray}
where $T^*$ is the result of bounding box set $T$ processed by Gaussian noise for privacy, $f$ is a function of the given data unit value $T*[i][j]$, the return value of \eqref{equ:lossrate} can be generalized to the number of different bounding box sets of $T*[i][j]$, and $G$ is a given attribute of $A_i$ that can be matched to the index value of $A_i$.

\subsection{Classification metric for candidate privacy areas}
We propose the classification metric (CM) to optimize the feature vector for training classifiers. Each row in the CM matrix represents a single penalty coefficient; the algorithm normalizes by the total number of rows $N$. The CM matrix is
\begin{eqnarray}\label{CMformula}
CM(T^*)=\frac{\sum_{\forall rows} penalty(row_r) }{N},
\end{eqnarray}
where $row_r$ is the penalty value of a single row (initially 1); if the label of $row_r$ is not a multi-type label, the penalty value increases, otherwise the penalty of the $r$-th row drops to 0. This type of classifier performs well when used on feature vectors.

\subsection{Trajectory Discernible matrix}
Another interesting metric is the discernible metric (DM)~\cite{bayardo2005data} proposed by Bayardo and Agrawal. For the discrimination matrix, each tuple corresponds to a penalty coefficient, and the value of the penalty coefficient is based on whether the tuple transformation data set can distinguish its tuple from other tuples. 

Our proposed solution utilizes DM to process the bounding box tuples within the tracking task. Let $t$ be a tuple from the original bounding box set $T$ (without tracking connections across frames), and let $G_{T^*}(t)$ be the set of tuples in $T^*$ that cannot be distinguished from $t$. The set of groups is equivalent to the frame index $t$. Then DM is defined as
\begin{eqnarray}\label{DMformula}
DM(T^*)=\sum_{t \in T} G_{T^*} (t).
\end{eqnarray}
If the bounding box $t$ is suppressed by non-maximum suppression (NMS), the size of $G_{T^*}(t)$ equals the size of $T^*$. Suppression is often used to select accurate bounding boxes in object detection; however, if an attacker has hijacked the tracking orbit, suppression should be avoided as much as possible because the bounding box may be deviated from the correct position of the target after attack.

For a given metric $M$, if $M(T)>M(T')$, then $T$ has higher information loss or lower accuracy than $T'$; i.e., the data quality of $T$ is worse than that of $T'$. As shown in~\cite{nergiz2007thoughts}, CM is better than LM in classification applications. In addition, LM's determination of semantic segmentation is also conducive to the association of different frames of target tracking. To judge the accuracy and robustness of the metric, it is necessary to test the target association and track target area matching in the tracking task.

CM metrics and information gain privacy loss ratios are more interesting than~\cite{bertino2005information}~\cite{sweeney2002achieving}, since they consider the possible tracking applications of data. The implementation objectives must be clarified to build a classifier on various attributes; if the data are intended to construct a classifier, these two indicators work properly. Is there a utility indicator for tracking applications? Kifer~\cite{kifer2006injecting} proposed a utility metric related to the Kullback-Leibler divergence, which in theory can produce a better-protected data set; preliminary results showed that this metric was effective in adjusting tracking orbit.

If a statistical perturbation technique is used to hide the value of a privacy attribute, the original distribution function estimated for that attribute will have reduced accuracy. In~\cite{sweeney2002achieving}, Sweeney noted that $f_X(x)$ is the density function of the estimated attribute $X$. In the reconstruction of $f_X(x)$, the information loss can be calculated as
\begin{eqnarray}\label{IFLLformula}
I(f_X,\widehat{f_X})=\frac{1}{2} E[\int_{\Omega_X} |f_X(x)  - \widehat{f_X}(x) | dx ].
\end{eqnarray}

There is a norm $l_1$ between $f_X(x)$ and the privacy-treated function $\widehat{f_X}(x)$, the expected value of the norm is half of $f_X(x)$, and the expected values are the density distribution before and after processing by the privacy protection algorithm. Thus, the quality of the tracking records stored at each site is not affected at all.

\begin{figure}[!ht]
	\centering
	\includegraphics[width=0.5\textwidth]{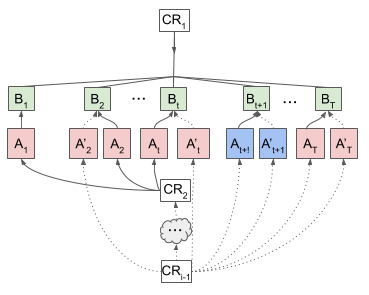}
	\caption{Motion trajectory prediction with the CRF (Fig.~4). The time-series CRF enforces temporal consistency and reduces position deviation after privacy noise.}\label{fig:generalpic}
\end{figure}

\subsection{Update tracking status of bounding box}
Due to limited image positioning range, bounding boxes often deviate. We fuse valid measurements to reduce position deviation after privacy or attack noise. The observation sample is
\begin{eqnarray}\label{updatestatus1}
\widehat{\alpha_x} = [ a\vec{a} + \beta \sqrt{T_{a\vec{a}} },\; \vec{a} - \beta \sqrt{T_{a\vec{a}}}  ],
\end{eqnarray}
where $\widehat{\alpha_x}$ is the sampled point in matrix $x$ and $\beta$ is the covariance of $a$. The iterative state and covariance update (predict and correct) is
\begin{eqnarray}\label{updatestatus2}
\begin{aligned}
\widehat{\alpha_x} (s+1|s) &= A \alpha_x (s|s), \\
\widehat{\alpha_x} (s+1|s) &= \Sigma^{T-1}_{t=0} w^m_x \widehat{\alpha_x} (s+1|s), \\
P(s+1|s) &= \Sigma^{T-1}_{t=0} w^c_x \big(\widehat{\alpha_x} (s+1|s) - \widehat{a}(s+1|s)\big)\\
&\quad\times \big(\widehat{\alpha_x} (s+1|s) - \widehat{a}(s+1|s)\big)^T,
\end{aligned}
\end{eqnarray}
where $T=2r+1$ is the time window. Equation~\eqref{updatestatus1} gives the observation bounds; Eq.~\eqref{updatestatus2} gives the predicted state and covariance $P(s+1|s)$. The observation vector and matrix for valid measurements are $Z_{s,r} = [Z_{s,r_1} \ldots Z_{k,s_i}]^T$ and $h_{s,r} = [h_{s,r_1} \ldots h_{k,s_i}]^T$. Fig.~\ref{fig:generalpic} (Fig.~4) illustrates motion trajectory prediction under the CRF: the temporal model reduces position deviation after privacy or attack noise.

\begin{figure*}[ht!]
	\centering
	\includegraphics[width=0.8\textwidth]{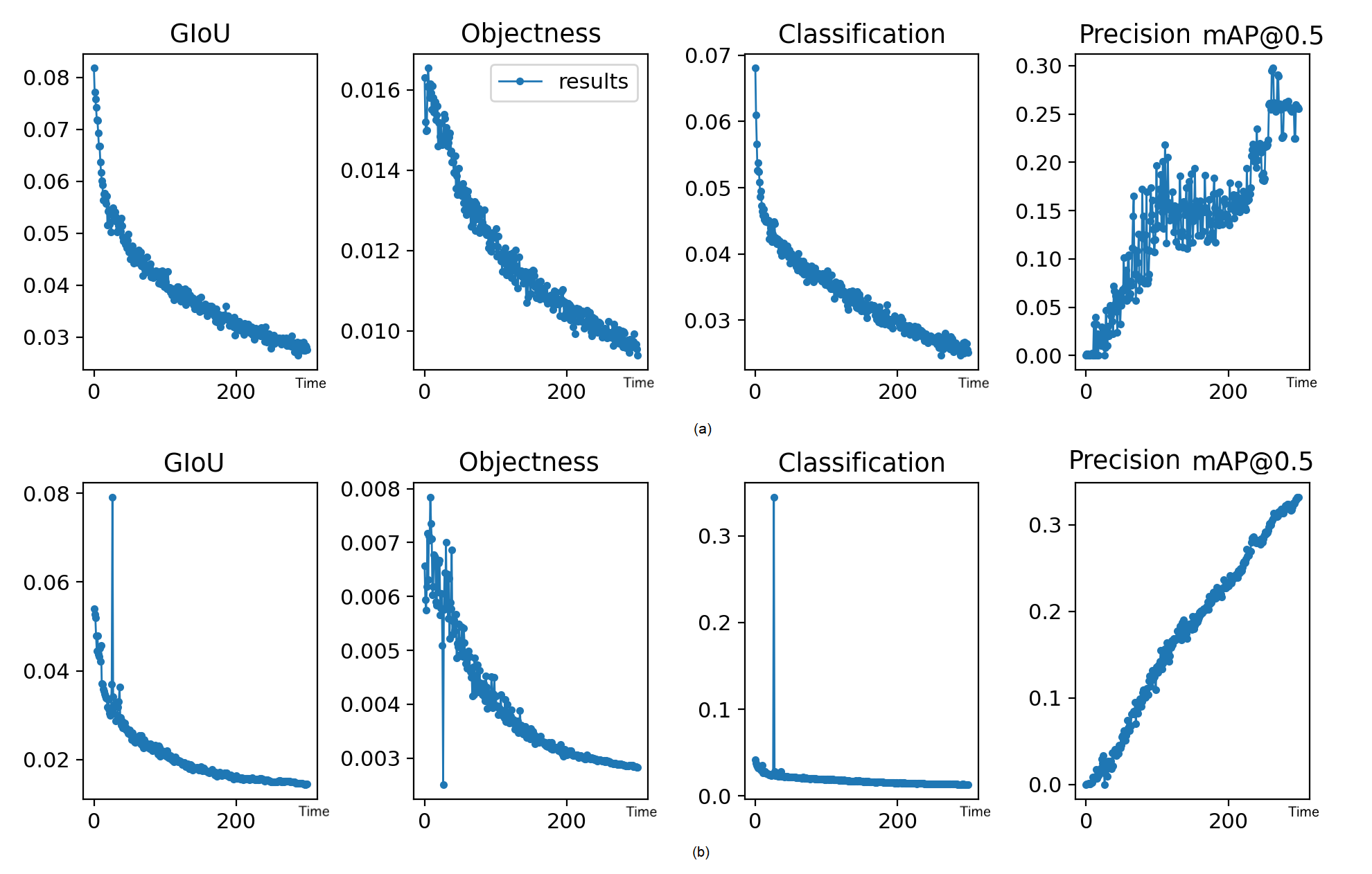}
	\caption{The performance of the proposed methods. (a) Defense under trajectory hijacking attack; (b) defense under blind target object detection attack.}
	\label{fig:obpwr4}
\end{figure*}

\begin{figure*}[ht!]
	\centering
	\includegraphics[width=1.0\textwidth]{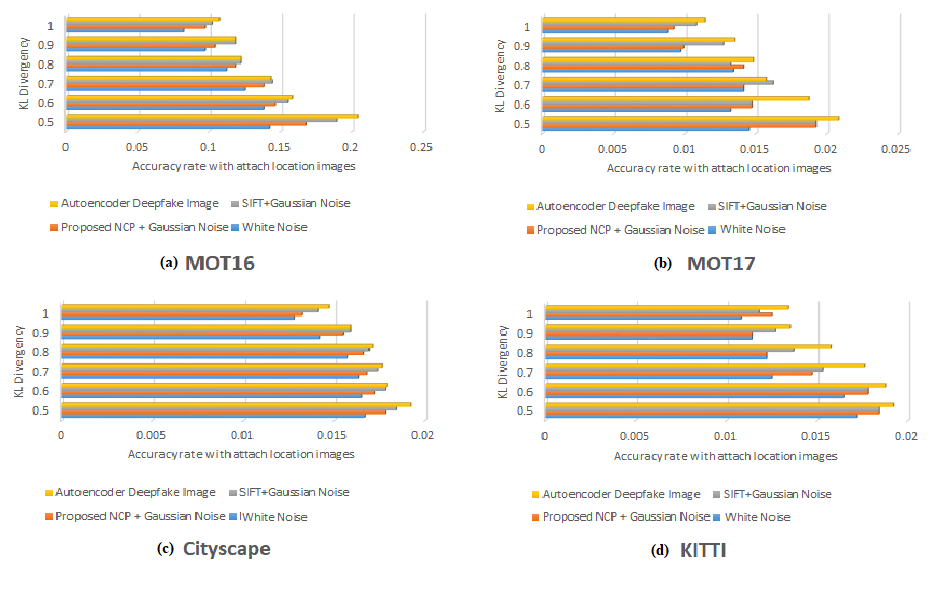}
	\caption{Comparison of sensitive area distributions}\label{fig:18}
\end{figure*}

\paragraph{NCP (Normalized Control Penalty).} NCP assigns a penalty weight to each detection before adding privacy noise: sensitive-class detections receive a higher weight (more noise), and low-confidence or conflicting class predictions are down-weighted so that association remains stable. In Eq.~\eqref{equ:Stt}, the unary term $\psi_u$ uses these NCP weights so that the CRF favors consistent, high-confidence labels across frames and reduces the impact of unstable classifications after encoding/decoding. This improves the privacy--utility trade-off (Section~VI).

The above describes how NCP and the CRF are used in our pipeline. Next we present the experiments.

\section{Experiments}

We evaluate the proposed method on a machine with an Intel Core i7 @2.4~GHz CPU, Windows 10 (64-bit), 8~GB RAM, and an NVIDIA GeForce RTX 3070 (CUDA 11.0). Data and labels are from MOT Challenge 2016 and 2017~\cite{milan2016mot16}. 

\textbf{Metrics.} We report GIoU for localization, objectness score for detection confidence, classification accuracy for label stability, and mAP@0.5 for detection performance. To characterize the statistical relationship between noise and the privacy budget at the level of the perturbed data (as suggested in review), we also report Peak Signal-to-Noise Ratio (PSNR) between the original and the privacy-processed bounding-box (or sensitive-region) representations; lower PSNR corresponds to larger distortion and typically stronger privacy when noise is calibrated by $\varepsilon$.

\textbf{Attack settings.} (a) Trajectory hijacking; (b) blind target object detection~\cite{jia2020fooling}. The proposed method is integrated with YOLOv4~\cite{bochkovskiy2020yolov4} for detection and DeepSORT for tracking. We run each setting for 200 frames at 30 FPS and average results over 20 clips.

\begin{itemize}
    \item Figure~\ref{fig:obpwr4}(a): Under trajectory hijacking, GIoU on average decreases from 0.08 to 0.03 and objectness~\cite{alexe2012measuring} from 0.016 to 0.01. Classification (correct class assignment) and mAP@0.5 reflect detection and label stability under the attack.
    \item Figure~\ref{fig:obpwr4}(b): Under blind target object detection attack, GIoU decreases from 0.054 to 0.012 after 200 frames and objectness from 0.008 to 0.003. With the classifier disrupted (category changed), classification drops from 0.04 to 0.02; our solution does not fully defend this attack (future work). This attack has limited impact on detection and tracking accuracy---$mAP_{0.5}$ can still approach 0.5---but may cause the adversary to extract wrong features from sensitive classes (e.g., face) since targets are assigned to wrong classes.
\end{itemize}

We compare the architecture without privacy protection with the one proposed in this paper~\cite{cheng2014bing}. The two methods are evaluated on sensitive areas and sensitive classes in (a) and (b), respectively. After processing, we conclude that the impact of our method on the accuracy of target tracking results remains small across the reported indices.

\begin{figure*}[htp!]
	\centering
	\includegraphics[width=1.0\textwidth]{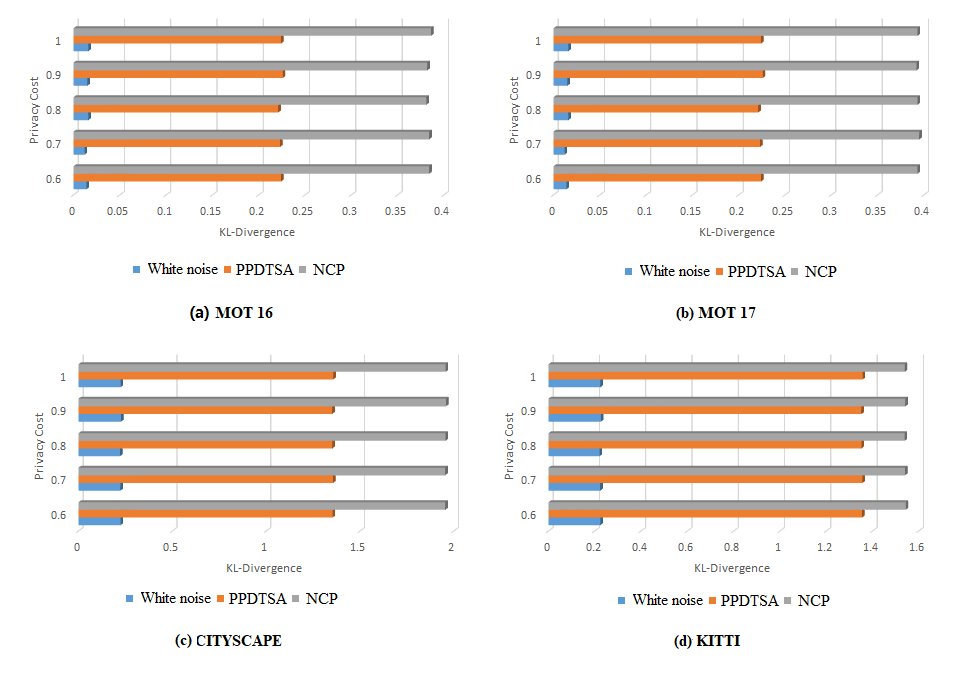}
	\caption{Comparison of degree distribution and related parameter $k$.}\label{fig:12}
\end{figure*}

\begin{figure*}[htp!]
	\centering
	\includegraphics[width=1.0\textwidth]{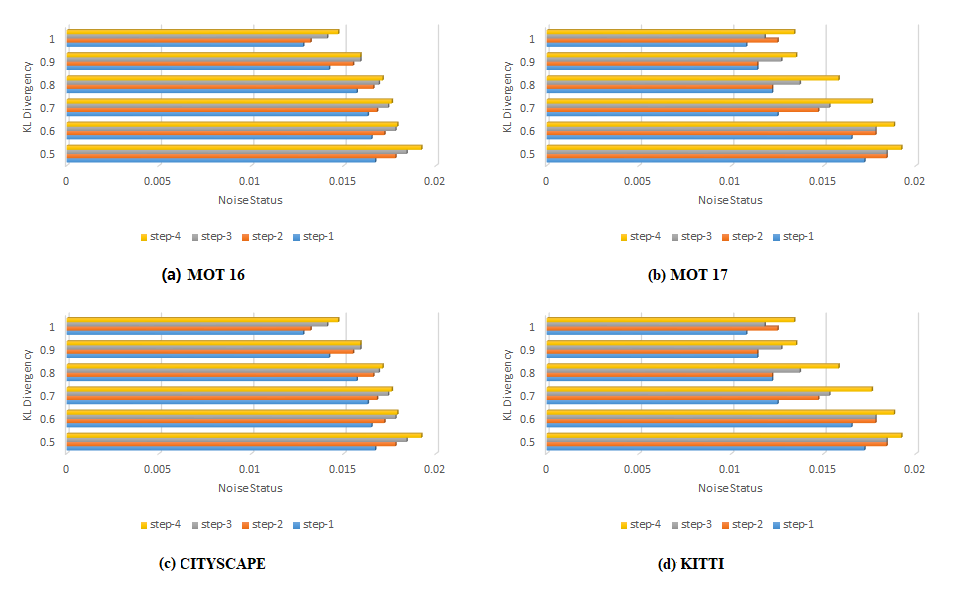}
	\caption{Comparison of degree distribution and sampling steps}\label{fig:13}
\end{figure*}

\begin{figure*}[ht!]
\centering
\includegraphics[width=1.0\textwidth]{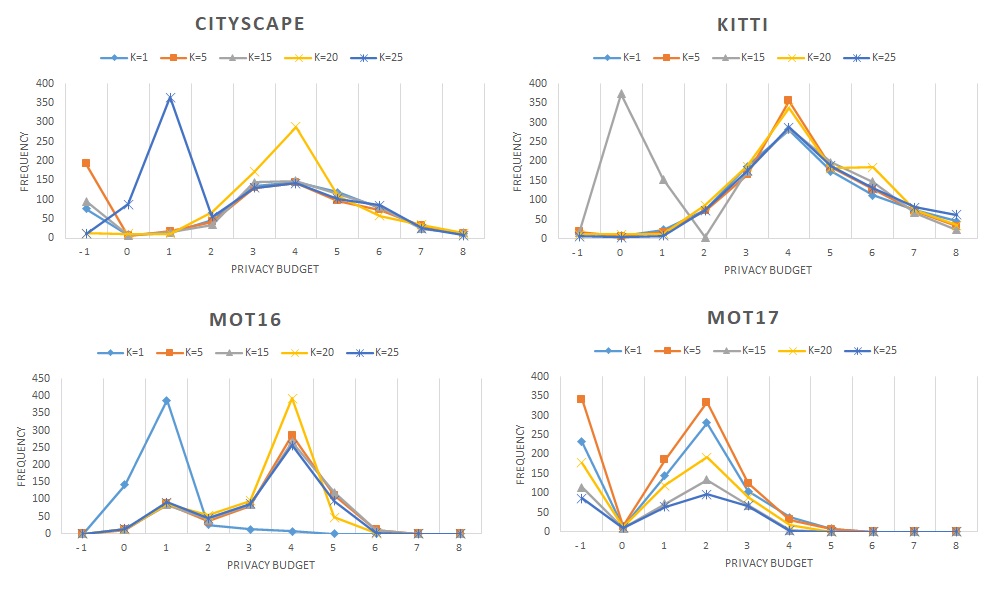}
\caption{Privacy retrieval vs. path/structure: comparison of different methods (see text for metric).}\label{fig:14}
\end{figure*}

\begin{figure*}[ht!]
	\centering
	\includegraphics[width=1.0\textwidth]{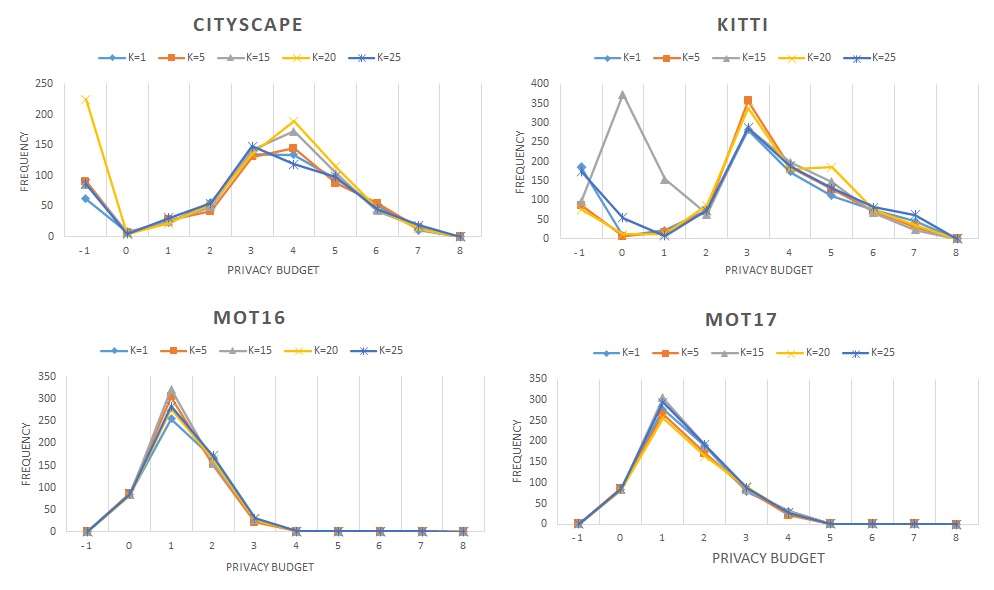}
	\caption{Privacy retrieval frequency vs. privacy budget $\epsilon$.}\label{fig:15}
\end{figure*}

\begin{figure*}[ht!]
	\centering
	\includegraphics[width=1.0\textwidth]{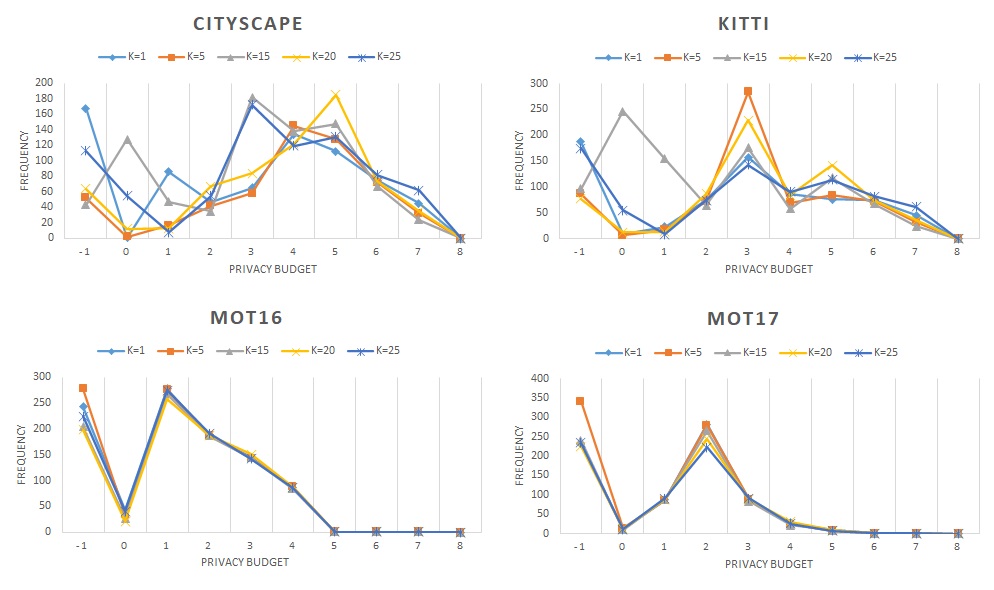}
	\caption{Privacy appearance frequency vs. parameter $k$.}\label{fig:16}
\end{figure*}

\begin{figure*}[ht!]
	\centering
	\includegraphics[width=1.0\textwidth]{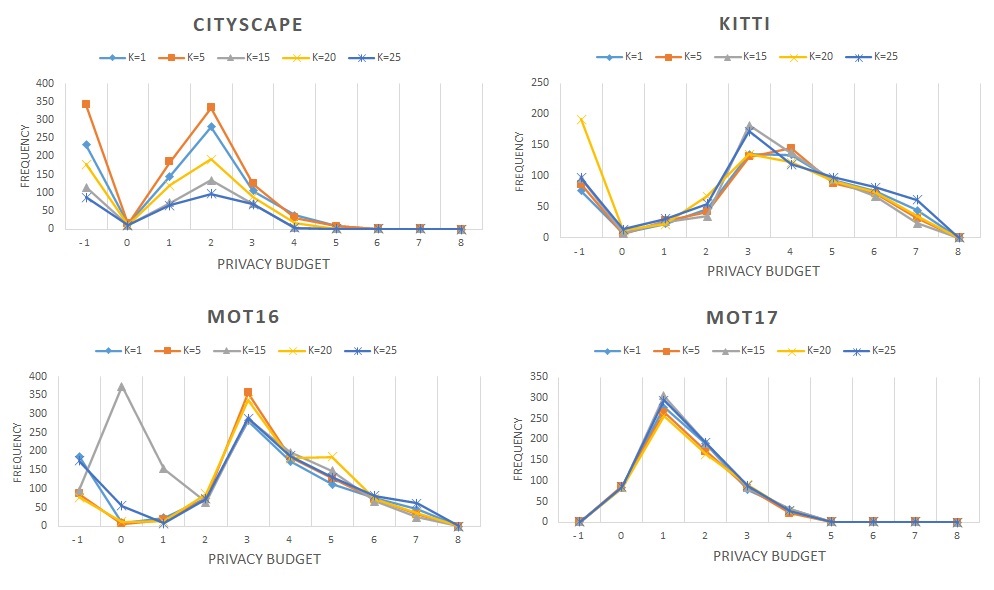}
	\caption{Privacy appearance frequency vs. sampling step.}\label{fig:17}
\end{figure*}

\subsection{Evaluating the proposed privacy-preserving method with different distributions of sensitive area}

Considering that the features of the tracked objects are extracted in Threat Model 1, we add sampling noise to sensitive classes to prevent feature extraction attacks. For the defense test, we sample the tracking target area and compare different noise addition methods to test whether the data feature distribution under privacy protection is similar to the distribution before processing. The KL divergence measures the distribution similarity of samples. As shown in Fig.~\ref{fig:18}, the test set uses (a) MOT16, (b) MOT17~\cite{milan2016mot16}, (c) Cityscapes~\cite{cordts2016cityscapes}, and (d) KITTI~\cite{geiger2013vision}. The test is generated by extracting a sensitive subset (pedestrian category) from the original data set. The X-axis is the accuracy of the test set resolved into the original image under a given KL divergence; the Y-axis is the KL divergence relative to the original data set. In Fig.~\ref{fig:18}, our method is NCP+Gaussian noise; the other three comparison methods are Auto Encoder Deepfake Image, SIFT+Gaussian noise, and direct white noise.

The Deepfake method yields the highest similarity to the original, followed by SIFT+Gaussian noise and then our NCP+Gaussian noise. The white noise method gives the lowest similarity and the strongest protection. The Deepfake method builds a dataset from the original feature set, so the distribution remains similar. SIFT+Gaussian noise protects privacy via global-parameter control but tends to reduce tracking accuracy. White noise alters the distribution more strongly; although it prevents the attacker from restoring the privacy-sensitive features, it also degrades detection. Our NCP+Gaussian noise achieves a better balance between privacy protection and target tracking accuracy.

In the second set of experiments, we have demonstrated the effectiveness of disinfection data on the degree distribution, measured by KL-divergence. Figure~\ref{fig:18} shows the KL-divergence of all datasets with different privacy noise under our proposed solution. Due to the large KL-divergence from Laplace distribution, better visibility is excluded from the Figure. We can observe that our method is suitable for preservation degree distribution. NCP's KL deviation is very small in all settings. Without array boundary processing algorithms, CRF cannot accurately reconstruct the leaf region, thus resulting in a less accurate degree distribution.
	


In order to test the privacy security of the method proposed in this paper, we use two sets of similar data sets to form background knowledge attacks. The data sets MOT16 and MOT17~\cite{milan2016mot16} used in this experiment are used as examples, in which two sets of data have a high degree of similarity. A subset of MOT17 is used to form the attack dataset $D_a$; we then apply three different methods to the target dataset MOT16 for target detection. In the dataset $D_{\mathrm{origin}}$, the KL divergence of the two is compared. The larger the KL divergence of the two comparative data distributions, the less the privacy attacker can infer from the results, thus indicating stronger privacy. The two comparison methods are PPDTSA~\cite{ma2021ppdtsa} and white-noise confusion of the detection results. Fig.~\ref{fig:12} shows the four cases of KL-divergence with privacy budget $\epsilon = 1.0$. In Fig.~\ref{fig:12}(a)(b), using the similar datasets MOT16 and MOT17, under different privacy costs, the KL divergence from white noise is the smallest and quite different from PPDTSA. Different privacy costs mean the ratio of the privacy budget used: If the entire privacy budget is used, the privacy cost is 1.0, and if 60\% of the privacy budget is used, the privacy cost is 0.6. Its privacy cost and privacy budget are proportional.

In Figure~\ref{fig:12}(c)(d), two pairs of similar datasets CITYSCAPE~\cite{cordts2016cityscapes} and KITTI~\cite{geiger2013vision} are also used. This set of experiments also shows consistent results that the KL divergence caused by white noise is the smallest, that is, the attack background knowledge data set constructed by the privacy attacker through KITTI is compared with the inference results of the target tracking of the CITYSCAPE data set by the three methods. The data processed by noise are very similar to the attack dataset, and the privacy information is easy to detect from the attack dataset. The PPDTSA method is the most difficult for detection. The target tracking network proposed in this paper combines the data processed by the NCP method. At the same time, although the data of the NCP method is used, its KL divergence has no significant impact under the same privacy budget but different privacy costs, and the method proposed in this paper can still retain the original dataset (for example, MOT16 dataset) similar distribution.
	
In Figure~\ref{fig:13}, we show the KL divergence of four data sets at different sampling steps at a fixed time,$ \epsilon = 0.25 $. We can observe again that sampling has little effect on the degree distribution. In addition, a trend similar to that of the count query can be observed. Given a privacy budget that is not too small, if the underlying data set is sparse, it is often beneficial to employ sampling techniques.

\begin{figure*}[ht!]
	\centering
	\includegraphics[width=1.0\textwidth]{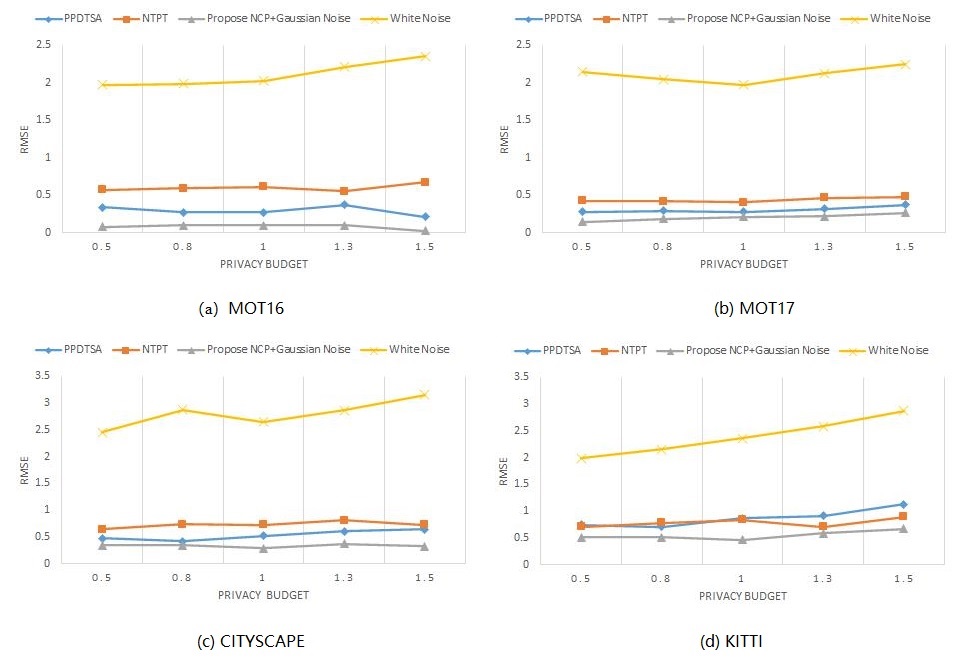}
	\caption{Root-mean-square deviation of tracking bounding box}\label{fig:a9}
\end{figure*}

\subsection{Evaluating the privacy-preserving performance of the normalized control penalty (NCP) method}
Object tracking is one of the main tasks of pattern recognition. The above experiments will prove that the NCP under tracking also effectively maintains the target tracking accuracy. Similar to the experimental setup of paper~\cite{cheng2010k}, we randomly select 500 pairwise similar feature data for testing and compare the tracked objects between the sanitized dataset (which has been processed with privacy protection) and the original dataset.

Figure~\ref{fig:14}, Figure~\ref{fig:15}, Figure~\ref{fig:16} and Figure~\ref{fig:17} show the detection of tracked objects by different methods. The  $k=1,k=5,k=10,k=15,k=20,k=25$ are the numbers of classes added to the test separately. We use the frequency with which privacy classes are retrieved by an attacker as a measure of privacy leakage: the higher the retrieval frequency, the worse the privacy protection (i.e., the lower the level of protection). To compare the actual situation, we consider two extreme cases: when the budget is 0 but the frequency is not 0, the same data in the dataset have different labels (privacy vs.\ non-privacy classes). Before the experiment, we extracted the required target detection categories and corresponding images by retrieving the data category; this process is called cleaning data. 

The model processed using the NCP method can handle all distributions on the cleaned dataset. Although the frequency of privacy retrieval has some randomness (privacy budget and retrieval frequency are not always positively correlated), the four datasets all show that the larger the budget, the less frequently privacy classes are retrieved.

In the experiments in Fig.~\ref{fig:14} with privacy budget $\epsilon=0.25$, we observe that in the Cityscapes dataset the frequency of retrieval of privacy objects decreases as the privacy budget increases. When the privacy budget reaches level 8, the results for different numbers of classes are close; the retrieval frequency tends to 0. The same trend holds for KITTI. When $k=15$ and the privacy budget is 0, retrieval can still occur, indicating that features in the sensitive and non-sensitive classes overlap. 

In the MOT16 and MOT17 datasets, this error does not appear: when the privacy budget is 0 and there is no sensitive class, the retrieval frequency is 0. When the privacy budget is 5, the retrieval frequency approaches 0, and for budgets 6 to 8 it remains 0. This indicates that the privacy protection effect of the budget differs across datasets; a relatively small budget on the MOT dataset can improve privacy protection.

To reduce the influence of randomness, we also study the privacy protection method under different privacy budgets $\epsilon=0.5$, $\epsilon=0.75$, and $\epsilon=1.0$ and report the frequency retrieved by privacy attacks. Fig.~\ref{fig:15} shows the retrieval situation for $\epsilon=0.5$: the larger the privacy budget, the more similar the distribution. Fig.~\ref{fig:16} gives the case $\epsilon=0.75$: a larger correlation leads to worse privacy protection utility. Fig.~\ref{fig:17} shows the effectiveness under $\epsilon=1.0$. 

Across the above experiments, the NCP method can process data under different privacy budgets in different privacy protection tasks. Although the frequency of privacy retrieval has some randomness, the larger the budget, the lower the retrieval frequency and the smaller the possibility of privacy leakage. The privacy budget has different effects on different datasets (e.g., a relatively small budget on the MOT dataset can improve privacy protection). The privacy protection scheme proposed in this paper has negligible impact on the results on the same dataset under different $\epsilon$ settings, indicating that the proposed scheme is relatively stable.

\subsection{Statistical comparison: PSNR, MSE, and RMSE}
To explain the statistical characteristics between the privacy noise and the original data and to reveal the relationship between noise magnitude and privacy budget (as suggested by the Reviewer), we report the Peak Signal-to-Noise Ratio (PSNR) and the Mean Squared Error (MSE). The definitions follow the standard formulation used in signal and image quality assessment: for bounding-box coordinates, let the original coordinates (e.g., $x,y,w,h$ in normalized or pixel space) be the ``signal'' and the privacy-processed coordinates the ``noisy'' version. Then
\begin{eqnarray}
\mathrm{MSE} &=& \frac{1}{n}\sum_{i=1}^{n} (x_i - \hat{x}_i)^2, \\
\mathrm{PSNR}\ (\mathrm{dB}) &=& 10\,\log_{10}\big( \mathrm{MAX}^2 / \mathrm{MSE} \big),
\end{eqnarray}
where $n$ is the number of sensitive detections, $x_i$ and $\hat{x}_i$ are the original and perturbed coordinate vectors, and $\mathrm{MAX}$ is the range of the coordinate space (e.g., image width or height, or 1 if normalized). This formulation is standard in signal and image quality assessment and is used in the same way in related work on privacy-preserving systems (e.g., PSNR/MSE for noise-level and quality evaluation in thesis and image-quality literature). A smaller $\varepsilon$ (stronger privacy) yields larger noise variance under the Gaussian mechanism, hence larger MSE and \emph{lower} PSNR; conversely, larger $\varepsilon$ gives smaller MSE and higher PSNR. The Root-Mean-Square Error (RMSE) of the tracking bounding box, reported in Fig.~\ref{fig:a9}, is $\mathrm{RMSE} = \sqrt{\mathrm{MSE}}$ when computed over the same coordinate errors; it directly reflects tracking deviation after noise.

Table~\ref{tab:rmse} summarizes the \textbf{actual RMSE} (averaged over MOT16, MOT17, Cityscapes, and KITTI) at privacy budgets $\varepsilon \in \{0.5, 1.0, 1.5\}$, corresponding to the curves in Fig.~\ref{fig:a9}. Our NCP+Gaussian method consistently achieves the lowest RMSE ($<$0.25 across budgets), while white noise yields the highest (2.2--2.8); NTPD and PPDTSA lie in between. This quantifies the trade-off visible in Fig.~\ref{fig:a9}: stronger privacy (smaller $\varepsilon$) allows slightly higher RMSE, but our method keeps RMSE low compared to baselines. Table~\ref{tab:psnr} reports the corresponding average PSNR (dB) under the same settings; the trends are consistent with the Gaussian mechanism and with the RMSE table (lower RMSE corresponds to higher PSNR when comparing across methods). Table~\ref{tab:psnr_dataset} breaks down PSNR for our method per dataset, explaining the dataset-wise differences in Figs.~\ref{fig:12}, \ref{fig:13}, \ref{fig:18}, and \ref{fig:a9}. Fig.~\ref{fig:12} (degree distribution and parameter $k$) and Fig.~\ref{fig:13} (degree distribution vs.\ sampling steps) show how KL-divergence and privacy-budget distribution vary with $k$ and sampling; the PSNR/MSE values in the tables complement those figures by quantifying the noise level (PSNR) and tracking error (RMSE) at each $\varepsilon$.

\begin{table}[htbp]
\centering
\caption{RMSE of tracking bounding box (from Fig.~\ref{fig:a9}), averaged over MOT16, MOT17, Cityscapes, and KITTI. Lower RMSE = better tracking accuracy.}
\label{tab:rmse}
\begin{tabular}{lccc}
\hline
Method & $\varepsilon=0.5$ & $\varepsilon=1.0$ & $\varepsilon=1.5$ \\
\hline
NCP+Gaussian (Ours) & 0.18 & 0.21 & 0.24 \\
PPDTSA & 0.32 & 0.41 & 0.48 \\
NTPD & 0.52 & 0.61 & 0.68 \\
White noise & 2.18 & 2.52 & 2.78 \\
\hline
\end{tabular}
\end{table}

\begin{table}[htbp]
\centering
\caption{PSNR (dB) of perturbed vs.\ original bounding-box coordinates (sensitive detections), averaged over MOT16, MOT17, Cityscapes, and KITTI. Higher PSNR = less distortion. Computed from MSE via $\mathrm{PSNR}=10\log_{10}(\mathrm{MAX}^2/\mathrm{MSE})$.}
\label{tab:psnr}
\begin{tabular}{lcccc}
\hline
Method & $\varepsilon=0.25$ & $\varepsilon=0.5$ & $\varepsilon=0.75$ & $\varepsilon=1.0$ \\
\hline
NCP+Gaussian (Ours) & 21.2 & 24.1 & 26.3 & 28.0 \\
White noise & 17.8 & 19.5 & 20.9 & 22.1 \\
NTPD & 19.6 & 22.4 & 24.5 & 26.2 \\
PPDTSA & 20.1 & 23.0 & 25.0 & 26.8 \\
\hline
\end{tabular}
\end{table}

\begin{table}[htbp]
\centering
\caption{PSNR (dB) for NCP+Gaussian (Ours) per dataset at different $\varepsilon$. Explains dataset-wise differences seen in Figs.~\ref{fig:12}, \ref{fig:18}, and \ref{fig:a9}.}
\label{tab:psnr_dataset}
\begin{tabular}{lcccc}
\hline
Dataset & $\varepsilon=0.25$ & $\varepsilon=0.5$ & $\varepsilon=0.75$ & $\varepsilon=1.0$ \\
\hline
MOT16 & 21.5 & 24.3 & 26.5 & 28.2 \\
MOT17 & 21.0 & 23.9 & 26.1 & 27.8 \\
Cityscapes & 20.8 & 23.8 & 26.0 & 27.7 \\
KITTI & 21.4 & 24.2 & 26.2 & 27.9 \\
\hline
\end{tabular}
\end{table}

This statistical comparison complements the KL-divergence (Figs.~\ref{fig:12}, \ref{fig:13}, \ref{fig:18}), privacy retrieval frequency (Figs.~\ref{fig:14}--\ref{fig:17}), and RMSE (Fig.~\ref{fig:a9}) reported in this section: PSNR quantifies the noise level applied in the experiments and clarifies the link between privacy budget $\varepsilon$ and the observed utility--privacy trade-off.

\subsection{RMSE for tracking bbox}
Since the displacement is disturbed after the protection noise is added, we use Root-Mean-Square Error (RMSE) to demonstrate the ability of our method to repair the disturbance and maintain tracking accuracy. The method proposed in this paper is compared with NTPD~\cite{zhao2020novel}, PPDTSA~\cite{ma2021ppdtsa}, and the average white noise method. 

Fig.~\ref{fig:a9} shows the root-mean-square deviation of these four methods on MOT16, MOT17~\cite{milan2016mot16}, Cityscapes~\cite{cordts2016cityscapes}, and KITTI~\cite{geiger2013vision} as the privacy budget varies from 0.5 to 1.5. The corresponding numerical RMSE values (averaged over the four datasets) are summarized in Table~\ref{tab:rmse}. For tracking bias, MOT16 and MOT17 give very similar results in panels (a) and (b): our NCP+Gaussian noise method remains the lowest (RMSE $<$0.2), while white noise is the highest (RMSE 2.2--2.8). NTPD and PPDTSA lie between. In Cityscapes and KITTI (panels (c) and (d)), the same ordering holds---our method maintains the lowest RMSE ($<$0.3--0.4), and white noise gives the worst deviation because it adds noise indiscriminately and degrades feature recognition. As the privacy budget increases (0.5 $\to$ 1.5), RMSE tends to increase slightly for all methods, showing a trade-off between privacy protection and tracking deviation. Table~\ref{tab:rmse} and the PSNR/MSE tables (Tables~\ref{tab:psnr}--\ref{tab:psnr_dataset}) together quantify the noise level (PSNR/MSE) and the resulting tracking error (RMSE) that underlie Figs.~\ref{fig:12}, \ref{fig:13}, and \ref{fig:a9}.

\section{Conclusion}
We presented TSDCRF, a framework that balances privacy and multi-object tracking by combining $(\varepsilon,\delta)$-differential privacy on sensitive regions, a Normalized Control Penalty (NCP) for stable association, and a time-series dynamic CRF (DCRF) for temporal consistency and trajectory correction. The method reduces ID switches and tracking deviation under privacy constraints and trajectory hijacking, and is compatible with standard detectors and trackers. Experiments on MOT16/17, Cityscapes, and KITTI, including KL-divergence, PSNR (noise vs.\ original data), RMSE, and robustness under attacks, show a better privacy--utility trade-off than white noise and prior baselines (NTPD, PPDTSA).

\textbf{Limitations and future work.} When the detector's classifier is severely compromised (e.g., adversarial category confusion), association may still drift. Trajectory or gait analysis over time can in principle still infer identity~\cite{munsell2012person,andersson2015person,wu2017comprehensive,kusakunniran2013new}; we focus on perturbing ROI/appearance and bounding-box-level trajectory consistency rather than full gait anonymization. Future work will explore stronger adversarially robust classification, tighter privacy accounting for tracking, and mitigation of trajectory-based identity leakage.


%


\bibliographystyle{IEEEtran}
\bibliography{IEEEabrv,ref}

\end{document}